\pdfoutput=1
\documentclass[11pt]{article}

\usepackage[preprint]{acl}
\usepackage{fontawesome}
\usepackage{times}
\usepackage{latexsym}
\usepackage{pifont}
\usepackage[T1]{fontenc}
\usepackage[utf8]{inputenc}

\usepackage{microtype}

\usepackage{inconsolata}

\usepackage{graphicx}

\title{Graph-Augmented Large Language Model Agents:\\ 
 Current Progress and Future Prospects \\
{\small \faGithub~\href{https://github.com/Shiy-Li/Awesome-Graph-augmented-LLM-Agent}{https://github.com/Shiy-Li/Awesome-Graph-augmented-LLM-Agent}}
}
\author{
  \textbf{Yixin Liu\textsuperscript{1}},
  \textbf{Guibin Zhang\textsuperscript{2}},
  \textbf{Kun Wang\textsuperscript{3}},
  \textbf{Shiyuan Li\textsuperscript{1}},
  \textbf{Shirui Pan\textsuperscript{1}}
\\
\\
  \textsuperscript{1}Griffith University, Gold Coast, QLD, 4215, Australia \\
  \textsuperscript{2}National University of Singapore, Singapore, 119077, Singapore \\
  \textsuperscript{3}Nanyang Technological University, Singapore, 639798, Singapore
\\
\small{
  \textbf{Correspondence:} \href{mailto:s.apn@griffith.edu.au}{s.pan@griffith.edu.au}
}
}

\begin{document}
\maketitle
\begin{abstract}
Autonomous agents based on large language models (LLMs) have demonstrated impressive capabilities in a wide range of applications, including web navigation, software development, and embodied control. While most LLMs are limited in several key agentic procedures, such as reliable planning, long-term memory, tool management, and multi-agent coordination, graphs can serve as a powerful auxiliary structure to enhance structure, continuity, and coordination in complex agent workflows. Given the rapid growth and fragmentation of research on \underline{G}raph-augmented \underline{L}LM \underline{A}gents (GLA), this paper offers a timely and comprehensive overview of recent advances and also highlights key directions for future work. Specifically, we categorize existing GLA methods by their primary functions in LLM agent systems, including planning, memory, and tool usage, and then analyze how graphs and graph learning algorithms contribute to each. For multi-agent systems, we further discuss how GLA solutions facilitate the orchestration, efficiency optimization, and trustworthiness of MAS. Finally, we highlight key future directions to advance this field, from improving structural adaptability to enabling unified, scalable, and multimodal GLA systems. We hope this paper can serve as a roadmap for future research on GLA and foster a deeper understanding of the role of graphs in LLM agent systems.

\end{abstract}

\section{Introduction}
The emergence of large language models (LLMs), such as GPT-4~\cite{achiam2023gpt}, Gemini~\cite{team2024gemini}, and DeepSeek~\cite{liu2024deepseek}, has brought about a paradigm shift in the design of autonomous agent systems~\cite{wang2024survey,guo2024large,xi2025rise}. By leveraging LLMs as central reasoning engines, a new generation of LLM-based autonomous agents (LLM agents for short) has demonstrated promising capabilities in understanding natural language instructions, executing multi-step tasks, and coordinating with external tools or services~\cite{zhang2025appagent,wu2024autogen,zhai2024fine}. The emergence of LLM agents enables them to tackle challenges once considered beyond the reach of artificial intelligence, opening doors to novel applications across a broad range of fields, such as software development~\cite{manish2024autonomous}, scientific research~\cite{m2024augmenting}, financial analysis~\cite{yu2024fincon}, etc.

\begin{figure}
    \centering
    \includegraphics[width=1.0\linewidth]{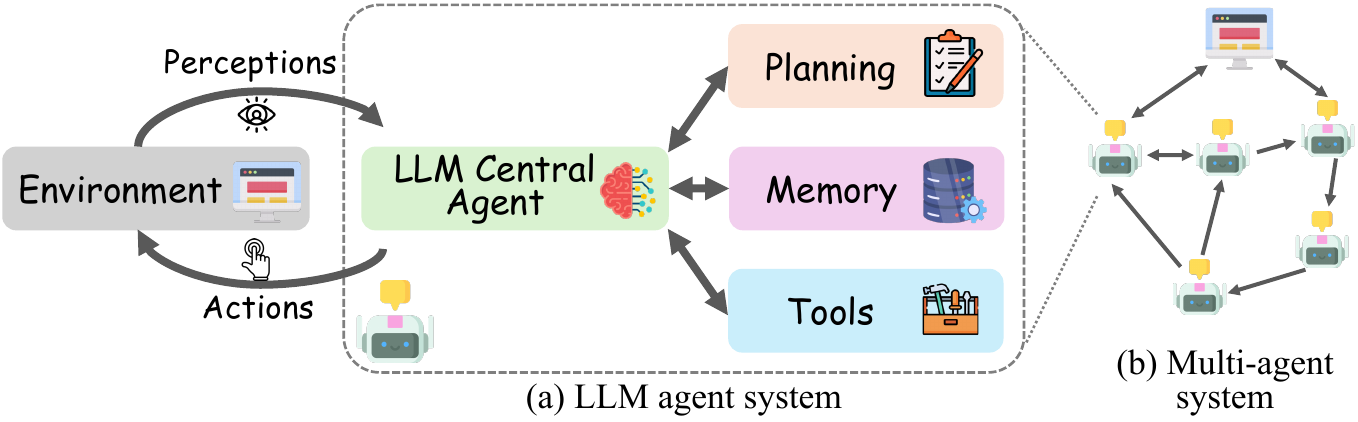}
    \caption{LLM agent framework and multi-agent system.}
    \label{fig:intro}
\end{figure}

Despite the impressive language understanding and reasoning capabilities of LLMs, they are insufficient on their own to support all key functionalities required in autonomous agent systems. As shown in Figure~\ref{fig:intro}, while LLMs serve as powerful cores for agent systems to interpret perceptions and make high-level decisions, they must be complemented with dedicated modules for \textbf{planning}, \textbf{memory}, and \textbf{tool using}, where LLMs may not perform reliably or efficiently. \textit{Firstly}, LLMs may perform unreliable task planning due to their tendency to hallucinate and limited understanding of multi-step dependencies~\cite{wu2024can}. \textit{Secondly}, LLMs struggle to maintain long-term memory efficiently in agent systems, primarily due to their stateless architecture and the constraints of a limited context window~\cite{fan2024survey}. \textit{Thirdly}, LLMs are inherently difficult to manage and invoke large toolsets due to their limited capacity for accurate selection, tool disambiguation, and consistent reasoning over unfamiliar or similar tools~\cite{liu2024toolnet}. \textit{Moreover}, when we extend a single agent to \textbf{multi-agent systems}, how to manage inter-agent communication and coordination remains an open question for LLMs~\cite{guo2024large}.

\begin{figure}
    \centering
    \includegraphics[width=1.0\linewidth]{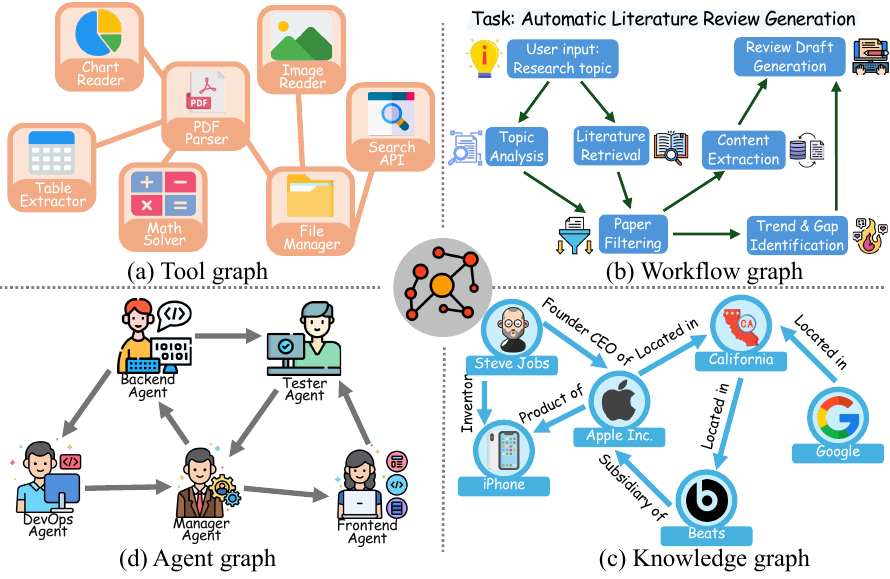}
    \caption{Different graphs in LLM agent systems.}
    \label{fig:intro_graphs}
\end{figure}

To address these limitations, researchers have increasingly turned to \textbf{graphs} as a complementary infrastructure to organize, enhance, and interpret the modules and execution flows of LLM agents, giving rise to an emerging research direction termed \underline{G}raph-augmented \underline{L}LM \underline{A}gents (GLA)~\cite{liu2024toolnet,zhuge2024gptswarm,zhang2025g}. Unlike the sequential language data on which LLMs are grounded, graphs are expressive and general-purpose data structures that naturally encode complex relationships among entities, tasks, tools, or agents. Within an LLM agent system, graphs can serve as tool managers (e.g., tool graphs, see Figure~\ref{fig:intro_graphs}(a)), task decomposition frameworks (e.g., task or workflow graphs, see Figure~\ref{fig:intro_graphs}(b)), external knowledge stores (e.g., knowledge graphs, see Figure~\ref{fig:intro_graphs}(c)), or communication infrastructures (e.g., agent coordination graphs, see Figure~\ref{fig:intro_graphs}(d)), and beyond~\cite{wu2024can,liu2024toolnet,zhang2025g,anokhin2024arigraph}. 

Compared to purely LLM-based solutions, GLA provides a range of benefits. 
\ding{182}~\textbf{Reliability}. Graphs ground the reasoning, memory, and knowledge of LLM agents on structured and factual data, reducing hallucinations and improving the reliability of agent systems~\cite{anokhin2024arigraph}.  
\ding{183}~\textbf{Efficiency}. Graphs support efficient information access and management by representing information in a compact, structured, and query-friendly form. Moreover, lightweight graph neural networks enable the training and deployment of auxiliary models over graph structures with minimal computational overhead~\cite{luo2025gfm}. 
\ding{184}~\textbf{Interpretability}. The explicit structure of graphs enhances explainability by revealing how information, tasks, and control signals propagate through the agent system, clarifying the rationale behind decisions and actions.
\ding{185}~\textbf{Flexibility}. Graphs promote modularity and reusability, allowing agents to generalize better across tasks by reusing graph-structured information, including but not limited to knowledge, memory, workflow, and communication patterns. 

\begin{figure*}
    \centering
    \includegraphics[width=1.0\textwidth]{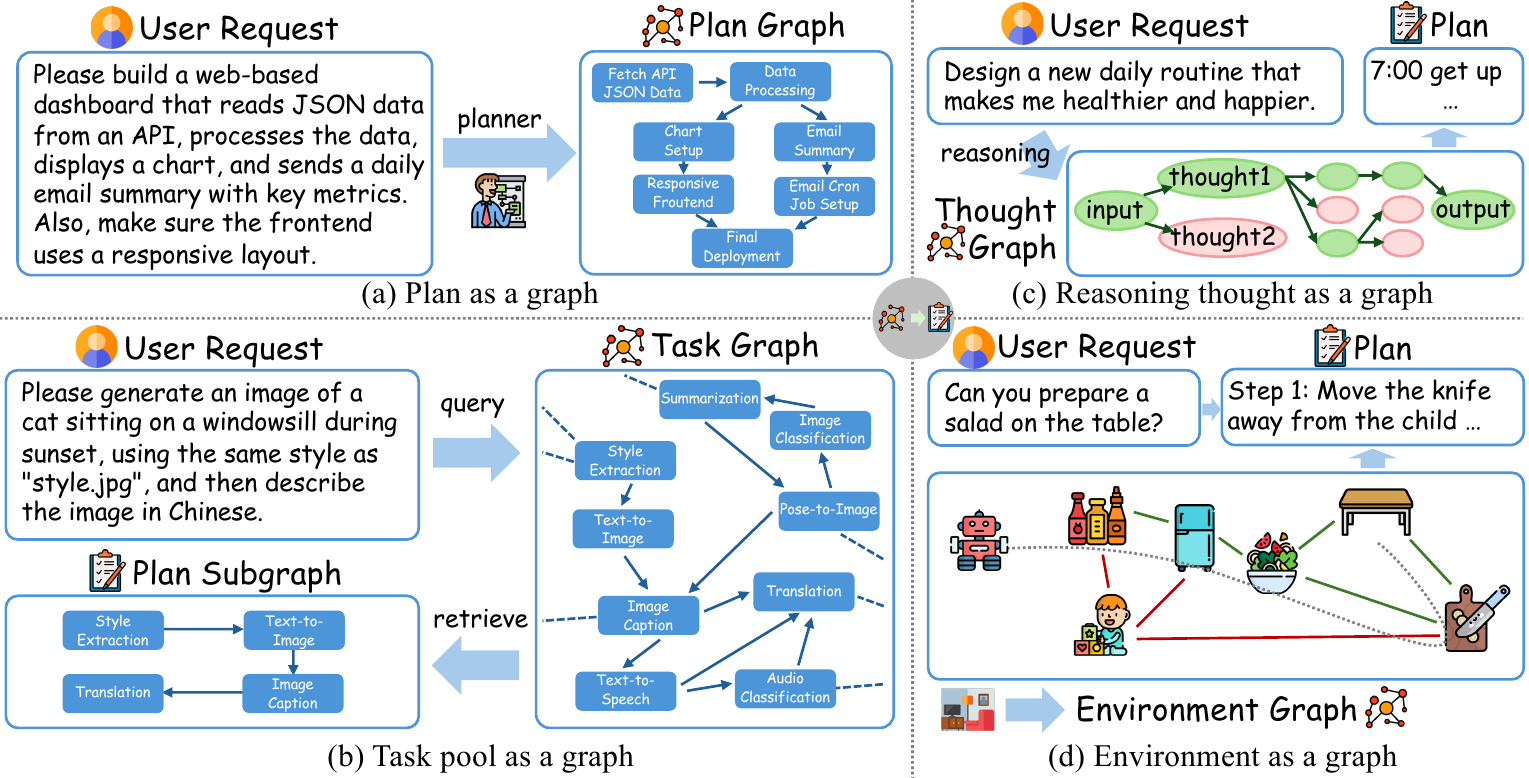}
    \caption{Graphs for planning in LLM agent system.}
    \label{fig:planning}
\end{figure*}

Despite GLA remaining an emerging research direction, there has been, to date, no comprehensive taxonomy or review article on this line of work. To fill the gap, this paper aims to provide a comprehensive overview of \textbf{existing studies} on GLA and offer insights into \textbf{future opportunities and challenges} in this emerging area. Specifically, we first discuss how graphs can augment \textit{\textbf{individual modules within the LLM agent framework}}, including planning and reasoning, memory and knowledge organization, and tool management. Then, we explore how graphs address unique challenges in \textit{\textbf{LLM-based multi-agent systems}}, particularly in designing collaborative workflows, ensuring safety, and optimizing efficiency. Finally, we highlight the key future directions, including dynamic and continual graph learning, unified graph abstractions for full-stack agent systems, multimodal graphs for multimodal agents, and large-scale multi-agent system simulation. We hope this article provides useful insights for ML and NLP researchers exploring structured representations in agent systems, graph learning practitioners seeking new applications, and system designers aiming to build more interpretable and collaborative LLM agent systems.

\section{Graph-Augmented LLM Agent System Framework}
% Generally speaking, 
Apart from the LLM central agent, the framework of an LLM agent system (as shown in Figure~\ref{fig:intro}(a)) is composed of three critical components: 
\ding{182}~\textbf{planning module}, responsible for decomposing tasks into actionable steps and reasoning for the task requirements; 
\ding{183}~\textbf{memory module}, which stores and retrieves contextual knowledge and/or historical memory to support task finishing; and
\ding{184}~\textbf{tool using module}, which enables the agent to interact with external APIs or environments to complete tasks beyond its internal capabilities. 
In each of these components, graphs play a critical role, taking on different forms and functions depending on the requirements of the specific module. In this section, we detail the ways in which graphs support planning, memory management, and tool management in LLM agent systems.

\subsection{Graphs for Agent Planning}

The planning module in LLM agent systems aims to decompose high-level goals into executable sub-tasks, determine the optimal action sequence, and guide the agent behavior in a structured and goal-directed manner. In this process, graphs play diverse roles to improve the reliability and quality of agent planning, assisting with sub-task sketching, sub-task organization, reasoning, and environment perception throughout the planning phase.

\subsubsection{\textbf{Plan as a graph}} 
In LLM agent systems, planning typically involves decomposing a complex high-level task into a group of manageable sub-tasks. During this decomposing process, the dependencies among sub-tasks can be represented as a graph, where nodes are sub-tasks and edges indicate their relationships, as shown in Figure~\ref{fig:planning}(a). The topological planning structure provides a clear and organized view of the task flow, enables the identification of reusable components, and supports efficient coordination of parallel or sequential execution. 

Guided by the idea of ``plan as a graph'', AFlow~\cite{zhang2024aflow} is a representative method that models agentic workflows as graphs where nodes represent LLM invocations and edges capture logical dependencies and execution flow. Based on the graph-based workflow modeling, AFlow leverages a Monte Carlo Tree Search strategy to automatically explore and optimize these workflows, reducing human effort while improving task performance. While AFlow shows promising improvements in automatically optimizing structured workflows for various LLM-based reasoning tasks, Wei et al.~\cite{wei2025vflow} further extends the graph-based workflow paradigm to agents for verilog code generation tasks. While AFlow constructs static agentic workflows through search-based optimization, AgentKit~\cite{wu2024agentkit} introduces a dynamic graph reasoning framework, where the execution graph evolves during interaction to support structured and context-aware decision making. Apart from optimizing and modeling agent planning, the plan graphs can also serve as structured supervision that teaches LLMs how to plan. By simulating different plan graphs, Plan-over-Graph~\cite{zhang2025plan} takes the synthetic plan graphs as explicit guidance to improve the parallel planning capabilities of LLMs through supervised fine-tuning and direct preference optimization. Due to their structured and interpretable nature, plan graphs have been recognized by a recent benchmark as a fundamental planning paradigm in agentic workflow generation~\cite{qiao2024benchmarking}.

\subsubsection{\textbf{Sub-task pool as a graph}} 
While directly converting a user request into a plan graph is a simple solution for task decomposition, it is hard to ensure that each node in the generated plan graph corresponds to a truly executable and meaningful sub-task. As several agentic platforms (e.g., HuggingGPT~\cite{shen2023hugginggpt}) provide a pool of pre-defined sub-task APIs, a more reliable solution is to ground the sub-tasks into a constrained task graph. In a task graph, each node corresponds to an available and executable sub-task, and each edge represents the dependency relationship between sub-tasks, typically indicating that the output of one task matches the input requirement of the next. Such an organization paradigm explicitly models the sub-task dependencies, allowing for more accurate and interpretable agent planning.

Building upon a sub-task pool graph constructed from HuggingGPT, Wu et al.~\cite{wu2024can} introduce a graph neural network (GNN)-based approach for agent planning. As illustrated in Figure~\ref{fig:planning}(b), given the user request as a query, the GNN model retrieves a plan subgraph composed of multiple pre-defined sub-tasks, representing the most suitable plan for the request. Empirical evidence shows that both training-free and training-based retrievers achieve reliable performance in generating executable plans, outperforming LLM-based planners that often suffer from hallucinations.

\subsubsection{\textbf{Reasoning thought as a graph}} 
Rather than directly generating plans from the user request, involving intermediate reasoning before planning has been proven effective in improving planning accuracy and reliability in LLM agent systems~\cite{huang2024understanding}.
Evolving from the Chain-of-Thought (CoT) paradigm~\cite{wei2022chain}, a recent research trend is to structure the reasoning flow as a graph. As demonstrated in Figure~\ref{fig:planning}(c), each intermediate thought can be represented as a node in a thought graph, where the edges encode logical connections or dependencies between reasoning steps. This graph-based organization of thoughts enables more flexible, interpretable, and self-refining reasoning, allowing LLM agents to refine, backtrack, or expand upon previous steps, leading to more coherent and reliable planning outcomes.

One pioneering approach is the Tree of Thoughts (ToT)~\cite{yao2023tree}, which enhances deliberate planning by exploring multiple reasoning paths, allowing language models to self-evaluate choices for complex problem-solving. 
While ToT provides a strong foundation for multi-path reasoning, it faces challenges in balancing factual accuracy and comprehensive logical optimization effectively. To address this limitation, RATT~\cite{zhang2025ratt} addresses the need for factually grounded planning by integrating retrieval-augmented generation to ensure both logical coherence and factual correctness at each step. 
Recognizing that tree structures can be restrictive for complex planning, Graph of Thought (GoT)~\cite{besta2024graph} models reasoning as an arbitrary graph. This allows for more adaptive planning through novel transformations like combining thoughts into synergistic outcomes. 
These graph-based planning frameworks have also been successfully adapted for domain-specific challenges. 
For instance, Thought Graph~\cite{hsu2024thought} applies these principles to planning in biological research, specifically for gene set analysis, achieving substantial improvements in uncovering semantic relationships between biological processesin uncovering semantic relationships between biological processes.
In addition, the goal-oriented thought graph~\cite{badagliacca2025graph} represents goals as explicit nodes, enabling transparent and verifiable reasoning processes.

\subsubsection{\textbf{Environment as a graph}} 
In addition to the user request, environmental perception is also vital for planning, as it informs the agent of contextual constraints and available actions. For example, for a robotic agent, modeling the environment directly determines the feasibility and efficiency of the generated action plan. In this context, graphs can be a powerful representation for modeling the environment. Figure~\ref{fig:planning}(d) provides an illustration of an environment graph, where entities in the real-world environment (i.e., a room) and their relationships (e.g., distance and interaction) are explicitly modeled, offering essential contextual information to support the decision-making and planning of the robotic agent.

Graphs can be used to describe the environments of various scenarios, from robotic agents to coding agents. For instance, Huang et al.~\cite{huang2025graphormer} propose to model safety constraints in robotic agent planning by constructing a dynamic spatio-semantic safety graph. The environment graph helps the LLM robotic agent to perform real-time hazard detection and adaptive task refinement during task planning. For coding agents, LocAgent~\cite{chen2025locagent} leverages code structure graphs to assist LLM agents in localizing and understanding buggy functions. By explicitly modeling function-level and semantic relationships across files, the graph empowers the agent to retrieve relevant contexts, reduce hallucination, and significantly improve bug localization accuracy.

\begin{figure}
    \centering
    \includegraphics[width=1.0\columnwidth]{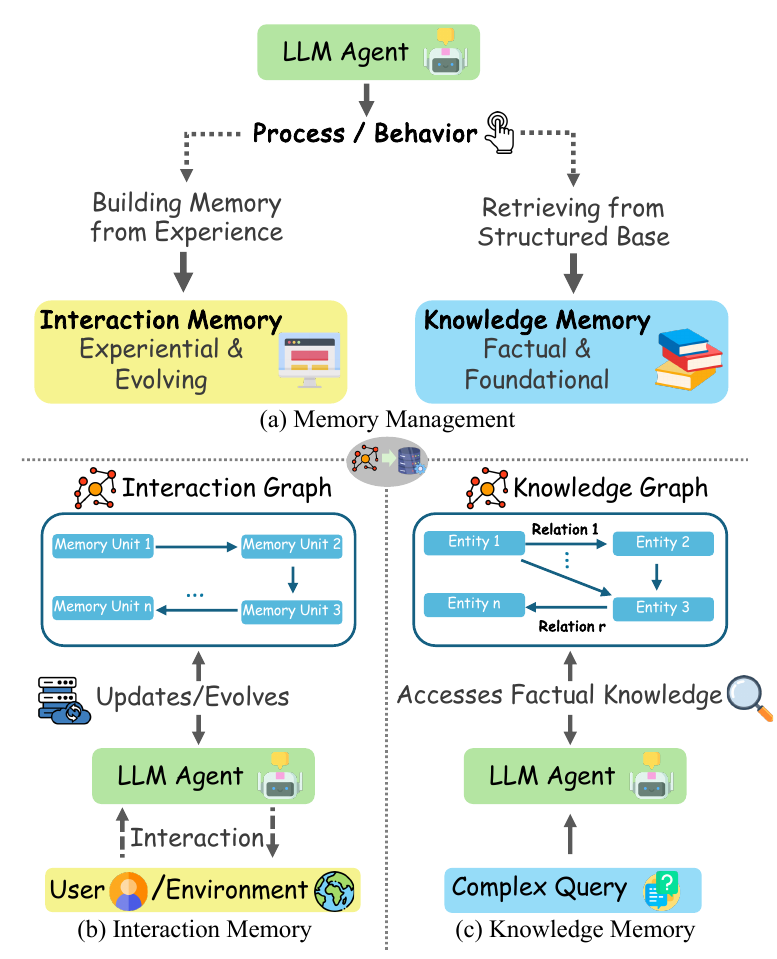}
    \caption{Graphs for memory management in LLM agent system.}
    \label{fig:memory}
\end{figure}

\subsection{Graphs for Agent Memory Management}
Effective memory management is essential for LLM agents to operate in complex, multi-step environments. As demonstrated in Figure~\ref{fig:memory}(a), agent memory can be categorized into two types: interaction memory and knowledge memory. Interaction memory captures and organizes the agent's experiences during environment or user interactions, preserving contextual continuity and enabling learning from past encounters. Knowledge memory, conversely, stores structured external information including facts, commonsense, and domain-specific knowledge that informs reasoning and decision-making processes. Graph structures offer a particularly powerful paradigm for organizing both memory types, as they naturally represent the interconnected nature of experiences and knowledge. By encoding information as nodes connected through meaningful relationships, graph-based memory systems enable efficient retrieval of contextually relevant information, support multi-hop reasoning across related concepts, and facilitate dynamic integration of new information into existing knowledge frameworks. This unified approach enhances the ability of agents to leverage past experiences, recognize patterns across interactions, and make informed decisions based on comprehensive contextual understanding.

\subsubsection{\textbf{Graph-organized interaction memory}} 
As LLM agents interact with environments and users over extended periods, they generate valuable experiential data that needs to be efficiently stored and retrieved. As shown in Figure~\ref{fig:memory}(b), graph-organized interaction memory represents these experiences as interconnected nodes and edges, where nodes typically represent interaction states, observations, or decisions, and edges capture temporal sequences or causal relationships. This structure enables agents to efficiently recall relevant past experiences, identify patterns across interactions, and leverage historical context for better decision-making. 

Recent advances in this area include A-MEM~\cite{xu2025mem}, which proposes an agentic memory system inspired by the Zettelkasten method to create interconnected knowledge networks through dynamic indexing and linking. When new memories are added, the system generates comprehensive notes with structured attributes including contextual descriptions, keywords, and tags, then analyzes historical memories to establish meaningful connections. This approach enables memory evolution, where new memories can trigger updates to existing representations, allowing the network to continuously refine its understanding. 
Similarly, AriGraph~\cite{anokhin2024arigraph} introduces a memory architecture that integrates semantic and episodic memories within a unified graph framework. As the agent interacts with its environment, each new observation generates an episodic vertex in the memory graph, while an LLM simultaneously extracts relationship triplets ($object_1$, $relation$, $object_2$) to update the semantic memory, with episodic edges connecting these memory types. This architecture is embedded within the Ariadne cognitive system, where a working memory holds recent observations and relevant knowledge, a planning module generates action plans, and a decision module selects actions for execution, with each observation triggering updates to the agent's world model.

\subsubsection{\textbf{Graph-organized knowledge memory}} 
Beyond managing interaction histories, LLM agents require access to structured external knowledge to support reasoning and decision-making. As demonstrated in Figure~\ref{fig:memory}(c), graph-organized knowledge memory represents domain knowledge as interconnected entities and relationships, enabling more nuanced understanding and inference capabilities. In these structures, nodes typically represent concepts, facts, or entities, while edges capture semantic relationships between them. This organization allows agents to navigate complex knowledge spaces, perform multi-hop reasoning, and integrate new information within existing knowledge frameworks. 

Several recent works have demonstrated the effectiveness of graph-organized knowledge memory for LLM agents. SLAK~\cite{zhou2024synergizing} constructs a location-based knowledge graph (LBKG) that integrates multi-sourced data from location-based social networks, enabling LLM agents to identify relevant meta-paths for socioeconomic prediction tasks. The framework introduces a cross-task communication mechanism that facilitates knowledge sharing at both agent and knowledge graph levels, significantly enhancing prediction accuracy through this synergistic design. In parallel, KG-Agent~\cite{jiang2024kg} proposes an autonomous framework that combines a multifunctional toolbox with a knowledge graph-based executor and dynamic memory system. This integration enables smaller language models to perform complex multi-hop reasoning through an iteration mechanism that autonomously selects appropriate tools and updates memory representations. By leveraging program language to formulate reasoning processes, KG-Agent demonstrates that structured knowledge representations can effectively compensate for model size limitations, outperforming larger models in both in-domain and out-domain question answering tasks.

\begin{figure}
    \centering
    \includegraphics[width=1.0\linewidth]{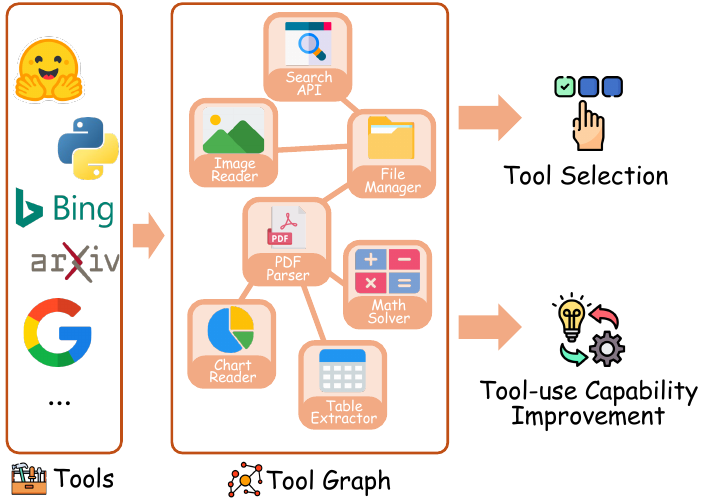}
    \caption{Graphs for tool management in LLM agent system.}
    \label{fig:tool}
\end{figure}

\subsection{Graphs for Tool Management}

The ability to use external tools is a fundamental capability for LLM agents to solve complex, real-world tasks. With the increasing number and variety of tools, effective tool management becomes critical to help agents select, coordinate, and leverage tools appropriately during complex tasks. To address the challenges of tool management, tool graphs provide a natural and structured way to represent the tool space. As shown in Figure~\ref{fig:tool}, in a tool graph, each node denotes an available tool for agents, and each edge models the functional dependencies and/or compatibility between tools. With the structured representation of the tool graph, we can not only perform accurate selection and retrieval of tools, but also enhance the tool-using capabilities of LLM agents.

\begin{figure*}
    \centering
    \includegraphics[width=1.0\textwidth]{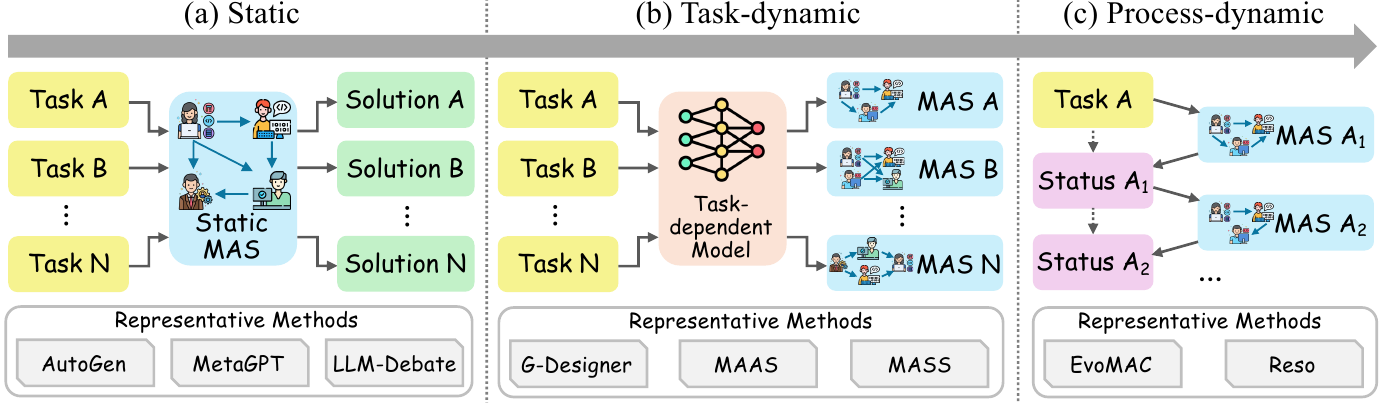}
    \caption{Graphs for MAS Orchestration.}
    \label{fig:mas_orche}
\end{figure*}

\subsubsection{\textbf{Tool graphs for tool selection}} 
As the number of available tools grows, selecting the appropriate tools given a complex user request becomes a non-trivial task for LLM agents. To address this challenge, graph-based representations provide an effective way for tool selection and retrieval. For instance, ControlLLM~\cite{liu2024controlllm} constructs a tool graph where nodes represent tools and resources, and edges model their input-output relationships. By searching over the tool graph, ControlLLM identifies executable toolchains that best satisfy the decomposed sub-tasks of the user request for LLM agents. SciToolAgent~\cite{chen2025scitoolagent} further advances this idea by leveraging a manually constructed scientific tool knowledge graph to guide LLMs in planning and executing multi-step toolchains across biology, chemistry, and materials science domains. ToolNet~\cite{liu2024toolnet} extends this line of work by organizing massive tools into a weighted directed graph, enabling LLMs to navigate the tool space efficiently with adaptive tool selection and dynamic updates based on prior usage. To sum up, the tool graph models both tool dependencies and transition preferences, enabling efficient and adaptive tool selection for LLM agents, especially from a large pool of tools. 

\subsubsection{\textbf{Tool graphs improve agent tool-use capability}} 
Despite the impressive capabilities across diverse tasks of LLMs, most LLMs are not inherently equipped with the ability to handle external tools properly. To enhance the tool-use capability of LLM agents, supervised fine-tuning with high-quality tool-interaction data has become an effective solution. To construct tool-interaction data, tool graphs offer a practical solution to sample related tool combinations as training examples, which helps improve the tool-calling capability of LLM agents. Following this idea, ToolFlow~\cite{wang2024toolflow} constructs a parameter-level tool graph based on semantic similarity between tool inputs and outputs, enabling the sampling of coherent tool combinations. The sampled tool subsets are then used to guide the generation of multi-turn dialogue plans, which serve as the fine-tuning supervision signals for LLMs to strengthen their tool-calling capability.

\section{Graph-Augmented LLM Multi-Agent Systems}

Recent studies demonstrate that organizing multiple LLM agents into multi-agent systems (MAS) can significantly surpass single-agent capabilities \cite{guo2024large,zhu2024survey,ning2024survey}. The inherent topological architecture of MAS naturally lends itself to graph-based modeling. Consequently, substantial research has focused on leveraging graph techniques to enhance MAS collaboration and complex task-solving abilities. This paper systematically investigates this through three dimensions: \ding{182} Fundamental designs for improving MAS performance via topology/workflow-driven graph concepts; \ding{183} Optimization of task-solving efficiency within MAS frameworks; and \ding{184} A principled synthesis of trustworthy MAS architectures to ensure reliability and deployability via graph-driven techniques.

\subsection{Graphs for MAS Orchestration}

A key distinction between MAS and single-agent setups lies in how MAS connect agents of diverse origins and capabilities to facilitate emergent collective intelligence. But what kind of structured language is suitable for describing, modeling, and managing such collective systems? Researchers from both the agent and graph communities have converged on \textit{graphs}—a data structure naturally suited for modeling connectivity among large-scale, heterogeneous entities. 
Intuitively, nodes in the graph represent individual agents, while edges capture their interactions. This modeling approach underpins early graph-based MAS systems such as GPTSwarm~\cite{zhuge2024gptswarm}, AgentPrune~\cite{zhang2024agentprune}, and MacNet~\cite{qian2024scaling-macnet}. 
However, agents often possess internal logic and modular structures. For instance, Anthropic's ``evaluator-optimizer'' pattern~\cite{2024} cannot be fully captured by simple node-edge representations. To accommodate such complexity, recent works introduce composite node abstractions to encode richer agentic structures, as seen in AFlow~\cite{zhang2024aflow} and MaAS~\cite{zhang2025maas}.
As a result, graphs have become a foundational tool for MAS modeling. Importantly, just as real-world graphs are dynamic and continuously evolving over time, graph-based MAS have also progressed from static topologies to task-adaptive, and more recently, process-adaptive structures. We will detail the progression in the following sections, and Figure~\ref{fig:mas_orche} provides an overview.

\subsubsection{\textbf{Static MAS Topology}} 
Early works did not explicitly adopt graph structures; for example, AutoGen~\cite{wu2024autogen} employed a chained layout, while DyLAN utilized a multi-layer perceptron-like architecture. A series of debate-based methods, such as ChatEval~\cite{chan2023chateval}, LLM-Debate~\cite{du2023improving}, and Persuasive-Debate~\cite{khan2024debating}, implicitly rely on graph-like structures to organize agent interactions.
More recent approaches model MAS orchestration explicitly as graphs. GPTSwarm~\cite{zhuge2024gptswarm} represents single-agent invocations as base nodes and composite behaviors (\textit{e.g.}, ReAct, CoT) as higher-level nodes. MacNet~\cite{qian2024scaling-macnet} explores a range of graph topologies—including tree, chain, star, random, and complete graphs—and evaluates their scalability as the number of agents grows.
AFlow~\cite{zhang2024aflow} introduces a two-level node abstraction: \textit{operators}, defined by a single LLM invocation with specific parameters (prompt, temperature, output format, \textit{etc.}), and \textit{nodes}, which group multiple operators into composite agentic units. These nodes are then connected to form agentic workflows. EvoFlow~\cite{zhang2025evoflow} follows a similar abstraction and applies a genetic algorithm to evolve a population of agentic workflows through query-specific evaluation.

\subsubsection{\textbf{Task-dynamic MAS Topology}} However, these topologies are task-independent, \textit{i.e.}, the same structure is applied regardless of domain or task complexity. G-Designer~\cite{zhang2025g} makes the first attempt at \emph{task-adaptive} topology orchestration, constructing MAS with varying graph complexity (\textit{e.g.}, node count, edge density) tailored to task difficulty. Specifically, it employs a variational graph auto-encoder to encode task-aware multi-agent graphs and generate topologies dynamically adjusted to task complexity. Meanwhile, EIB-learner~\cite{shen2025understanding} introduced a topological design method that integrates different connection patterns based on causal analysis results, better balancing error suppression and beneficial information propagation. In contrast, ARG-Designer~\cite{li2025assemble} models topological design as a graph generation task, creating customized topologies in a flexible and scalable manner based on autoregressive graph generation. Following this direction, \cite{zhang2025gnns-predict-mas} proposes using GNNs as performance predictors for MAS, accelerating MAS topology optimization. Similarly, MaAS~\cite{zhang2025maas} adopts a task-adaptive design paradigm based on an agentic supernet, further improving search efficiency and cost-effectiveness. MASS~\cite{zhou2025multi} provides a systematic analysis of how different topologies affect MAS performance.

\subsubsection{\textbf{Process-dynamic MAS Topology}} While these methods enable task-level adaptivity, they suffer from a key limitation: topologies are statically sampled prior to execution, lacking fine-grained adaptation and fault tolerance during runtime. ReSo~\cite{zhou2025reso} addresses this by decomposing the user query into a directed acyclic graph and performing dynamic planning and agent routing at each execution node, enabling more granular MAS evolution. Similarly, EvoMAC~\cite{hu2024self-evomac} adapts both topology and prompting based on environment feedback from each execution. AnyMAC~\cite{wang2025anymac} also performs step-wise topology planning guided by subtask progression.

\begin{figure}
    \centering
    \includegraphics[width=1.0\columnwidth]{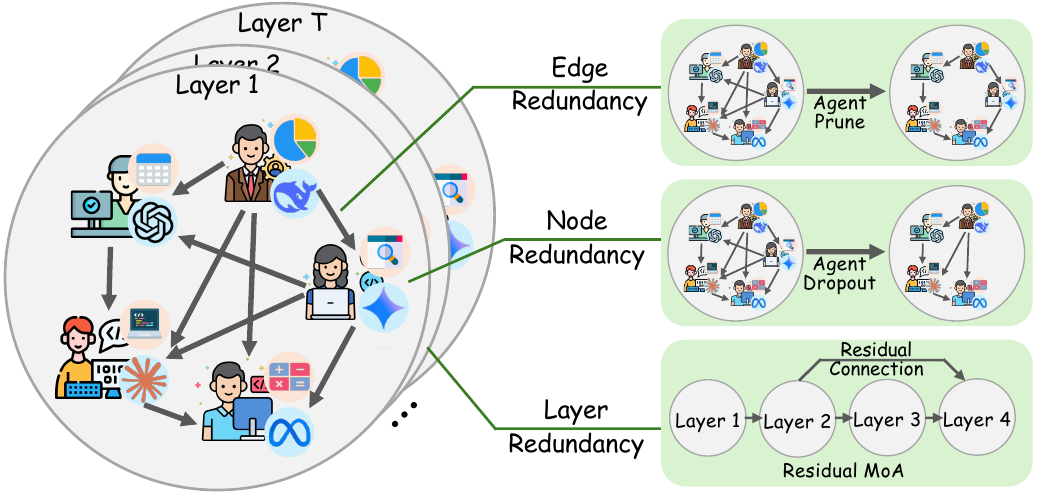}
    \caption{Graphs for MAS Efficiency.}
    \label{fig:mas_eff}
\end{figure}

\subsection{Graph for MAS Efficiency}

Just as real-world graph datasets often suffer from redundancy and noise, necessitating techniques such as graph structure learning~\cite{li2023gslb,zhu2021survey-gsl,li2025iceberg,li2023survey} and graph pruning~\cite{zhang2024gst,zhang2024mog,spielman2008graph-sparse-er} to refine the graph topology, graph-based MAS have likewise been observed to exhibit various forms of redundancy. For instance, many inter-agent communications are found to be ineffective or even detrimental~\cite{zhang2024agentprune,li2024improving-sparse-mad}; increasing the number of agents (nodes) does not always lead to performance gains~\cite{wang2025agentdropout}; and the marginal utility of additional multi-agent utterance rounds diminishes over time. Interestingly, these redundancy patterns parallel classical graph lightweighting phenomena: redundant inter-agent communication mirrors edge redundancy, excessive agents correspond to removable, unnecessary nodes, and diminishing returns from prolonged multi-agent interactions resemble the over-smoothing issue in GNNs, where deeper layers degrade performance. In what follows, we examine MAS efficiency issues from these three perspectives and discuss potential solutions inspired by graph-theoretic insights. A sketched illustration is provided in Figure~\ref{fig:mas_eff}.

\subsubsection{\textbf{Edge Redundancy in MAS}} A considerable number of MAS frameworks adopt predefined inter-agent communication paths (i.e., edges), as seen in DyLAN~\cite{liu2023dylan}, MachineSOM~\cite{zhang2023machinesom}, and ConsensusLLM~\cite{chen2023consensusllm}. Designing such topologies poses a critical challenge: how to strike a delicate balance between sufficient information exchange for task completion and minimizing token consumption. In practice, many predefined topologies are not readily conducive to effective multi-agent collaboration. For example, MacNet~\cite{qian2024scaling-macnet} evaluates various hand-crafted graph structures—MLP-like graphs, complete graphs, random graphs, stars, trees—but observes that performance does not consistently improve with increased edge density.
AgentPrune~\cite{zhang2024agentprune} is the first to formally define the \textit{Communication Redundancy} problem in MAS, demonstrating that many systems can achieve comparable performance even after pruning a subset of inter-agent communication edges. These removable edges are referred to as redundancies. Inspired by classical rank-based graph sparsification techniques, such as the low-rank sparsity loss in ProGNN~\cite{jin2020prognn}, AgentPrune models the MAS communication pipeline as a spatio-temporal graph and reformulates it as a learnable adjacency matrix. By optimizing a trainable graph mask, the method identifies and prunes redundant edges within the existing topology. Similarly, MAD~\cite{li2024improving-sparse-mad} demonstrates in multi-agent debate settings that sparse communication topologies can enhance both effectiveness and efficiency, while overly dense topologies often become burdensome. Addressing this issue, TalkHier~\cite{wang2025talk} proposes a well-structured, context-rich communication protocol that activates agent teams hierarchically, offering greater flexibility compared to approaches like AgentPrune and GPTSwarm, which rely on a fixed topology across all tasks. Overall, as task complexity increases and contextual inputs grow longer, edge redundancy remains a persistent challenge in MAS. Balancing communication token costs with system performance continues to be an open and important direction for future exploration.

\subsubsection{\textbf{Node Redundancy in MAS}} In conventional GNN research, it has been shown that randomly or strategically dropping edges or nodes can improve training efficiency and robustness without sacrificing, or even enhancing, performance, as observed in DropEdge~\cite{rong2019dropedge}, PTDNet~\cite{luo2021learning}, and DropGNN~\cite{papp2021dropgnn}. Interestingly, similar insights hold in MAS: AgentDropout~\cite{wang2025agentdropout} identifies underperforming or inactive agents and dynamically removes them, achieving node-level de-redundancy. Their approach also exhibits strong transferability, \textit{i.e.}, MAS pruned and optimized on one dataset generalize well to others.

\subsubsection{\textbf{Layer Redundancy in MAS}} Oversmoothing has long been a persistent challenge in the GNN community~\cite{chen2020measuring,cai2020note}: as the number of layers increases, quantitative performance often plateaus or even deteriorates. When mapped to MAS, a similar issue arises: system performance does not always improve with more rounds of multi-agent dialogue. This phenomenon was first observed in multi-agent debate frameworks: \cite{liang2023encouraging} found that while disagreement among agents decreases with more debate iterations, performance metrics do not consistently improve. Similarly, \cite{li2024improving-sparse-mad} showed that beyond five rounds of debate, additional iterations yield no performance gains. Although studies remain limited, recent work has begun to address MAS over-smoothing by drawing on GNN strategies. Residual Mixture-of-Agents (MoA)~\cite{xie2025rmoa} introduces residual connection modules that insert residual agents between generation rounds to compress historical information, enabling faster performance improvements in multi-turn MoA settings. Debate Only When Necessary (DOWN)~\cite{eo2025debate-down} introduces a simple yet effective \textit{Debate Engagement Check} to determine whether engaging in a debate is necessary, thus reducing unnecessary communication overhead.

\subsection{Graphs for Trustworthy MAS}

Trustworthiness is a fundamental requirement for ensuring the operational reliability of agentic systems~\cite{chen2025survey,wang2025comprehensive,yu2025survey,liu2025compromising,wang2025manipulating,liu2025agentsafe}. The unique topological structure of MAS has motivated research into how different network configurations affect the propagation of hazardous information. For instance, {NetSafe}~\cite{yu2024netsafe} and {ARGUS}~\cite{li2025goal} systematically investigated the flow of biases and harmful content through MAS architectures, while {AgentSafe}~\cite{mao2025agentsafe} further examined their potential impacts on memory subsystems.  Building on these foundations, {G-Safeguard}~\cite{wang2025g} leverages the inductive bias of graph neural networks \cite{wu2020comprehensive} to model threat propagation in MAS, enabling the prediction and detection of malicious nodes without requiring explicit predictor training. Complementing these approaches, {FlowReasoner}~\cite{liuyue_FlowReasoner} introduces a reasoning-based meta-agent to automate the design optimization of query-level MAS architectures.

Complementing technical research, several benchmarks have emerged to address agent safety concerns. {Agent-SafetyBench}~\cite{zhang2024agent} systematically investigates safety issues in simulated environments, while {AgentAuditor} \cite{luo2025agentauditor} provides formal safety evaluation frameworks for monitoring agent behavior throughout operational processes.

\section{Conclusion and Future Directions}
Despite recent progress in GLA, this research area remains in its early stages of development. To guide future exploration, we summarize several promising research opportunities below.

\subsection{Dynamic and Continual Graph Learning for Agent Systems}
Most current GLA systems rely on static or session-specific graph structures, which are constructed separately for each task and remain fixed throughout the execution process of agents. Nevertheless, real-world scenarios often involve dynamic environments and shifting task requirements, which demand continual updates to graph structures and graph learning modules. To adapt to evolving tasks and dynamic environments, a promising future direction is to develop dynamic and continual graph learning frameworks, where graph structures and representations evolve alongside agent interactions and environmental feedback. With the ability to incrementally update task plans, memory structures, and tool-use pathways based on new experiences, such systems can enable lifelong learning and persistent improvement, thereby enhancing agent adaptability in complex scenarios and bringing them closer to human-like learning behaviors.

\subsection{Unified Graph Abstractions for Full-Stack Agent Systems} 

In the existing methods for LLM agents, the graphs and graph learning modules are designed independently for individual modules, such as planning, memory, tool orchestration, or agent collaboration. However, full-stack agent intelligence often requires tight integration across these components to support coherent reasoning, modular reuse, and efficient adaptation. To this end, a promising path forward is to construct unified graph abstractions that holistically represent agent knowledge, workflows, and interactions across the full stack. To reach the goal, graph foundation models emerge as a compelling option: pretrained on large-scale agent-related graph data, they offer adaptable and reusable representations that can seamlessly integrate into various agent modules. A unified graph abstraction can help agents perform more consistent reasoning, facilitate information sharing across modules, and ultimately enhance generalization and scalability in complex, multi-task environments.

\subsection{Multimodal Graphs for Multimodal Agents}

As LLM agents are increasingly deployed in multimodal settings involving language, vision, audio, and action, a major challenge arises in organizing and reasoning over heterogeneous sensory inputs and outputs. Existing graph-based solutions for agents typically focus on textual or symbolic domains, lacking the flexibility to represent cross-modal relationships and temporal dynamics. To address this, future work can explore the development of multimodal graphs that unify diverse modalities within a coherent graph representation, where nodes correspond to visual objects, speech segments, text entities, or actions, and edges capture semantic, spatial, or temporal dependencies. Such graphs would enable agents to perform structured, multi-hop reasoning across modalities, facilitate modality alignment and fusion, and support tasks like grounded planning, embodied interaction, or video question answering.

\subsection{Graphs for Trustworthy Multi-Agent Systems}

The proliferation of multi-agent systems has led to a heightened focus on their trustworthiness, including crucial aspects such as security, fairness, and privacy. It is our contention that a graph modeling perspective can provide substantial support for enhancing these properties.
Taking federated multi-agent systems for example, where agents from disparate entities, each potentially possessing private data, must cooperate, ensuring that sensitive information is not divulged during inter-agent communication is critical. Graph-based methodologies present a promising avenue for resolving this issue. For instance, techniques like node decomposition combined with homomorphic encryption can be conceptualized within a graph framework to facilitate secure information exchange between agents. 
Furthermore, the inherent openness of many multi-agent systems renders them susceptible to security threats, such as the infiltration of malicious agents intent on executing attacks like prompt injection or memory poisoning. To fortify system security, inspiration can be drawn from the domain of graph learning. Methodologies for detecting malicious agents can be developed by modeling the system as a dynamic interaction graph and identifying anomalous patterns at the node, edge, or even subgraph level. Finally, the principle of fairness is indispensable for the long-term stability and effectiveness of multi-agent systems, especially those involving resource allocation or collaborative task execution. Graph theory provides a formal language to describe and analyze fairness in these contexts, fostering mechanisms that not only optimize for system efficiency but also ensure that outcomes are equitable for all participating agents, preventing issues like resource starvation and promoting sustained cooperation.

\subsection{Graphs for Large-Scale Multi-Agent System Simulation}

Simulating large-scale multi-agent systems has substantial real-world value, enabling researchers to explore emergent behaviors, evaluate coordination strategies, and stress-test agent architectures in complex settings such as smart cities, supply chains, and social platforms. However, most existing graph-based MAS approaches are confined to small-scale scenarios, typically modeling only a few dozen agents, and fall short in capturing the complexity and scale of population-level interactions. Scaling up introduces significant challenges, including communication overload, decentralized control, and the need for dynamic topology adaptation as agents and tasks evolve over time. To tackle these issues, future work can explore large-scale graph learning algorithms that support efficient representation, inference, and adaptation over massive, evolving agent interaction graphs. Such methods would enable more realistic MAS simulations and support the deployment of large-scale agent systems in diverse domains such as social behavior modeling, industrial robotics, autonomous traffic control, and multi-agent financial systems.

\bibliography{ref}

\begin{thebibliography}{93}
\providecommand{\natexlab}[1]{#1}

\bibitem[{Achiam et~al.(2023)Achiam, Adler, Agarwal, Ahmad, Akkaya, Aleman, Almeida, Altenschmidt, Altman, Anadkat et~al.}]{achiam2023gpt}
Josh Achiam, Steven Adler, Sandhini Agarwal, Lama Ahmad, Ilge Akkaya, Florencia~Leoni Aleman, Diogo Almeida, Janko Altenschmidt, Sam Altman, Shyamal Anadkat, and 1 others. 2023.
\newblock {GPT-4} technical report.
\newblock \emph{arXiv preprint arXiv:2303.08774}.

\bibitem[{Anokhin et~al.(2024)Anokhin, Semenov, Sorokin, Evseev, Burtsev, and Burnaev}]{anokhin2024arigraph}
Petr Anokhin, Nikita Semenov, Artyom Sorokin, Dmitry Evseev, Mikhail Burtsev, and Evgeny Burnaev. 2024.
\newblock Arigraph: Learning knowledge graph world models with episodic memory for llm agents.
\newblock \emph{arXiv preprint arXiv:2407.04363}.

\bibitem[{Anthropic(2024)}]{2024}
Anthropic. 2024.
\newblock \href {https://www.anthropic.com/engineering/building-effective-agents} {Building effective ai agents}.

\bibitem[{Badagliacca et~al.(2025)Badagliacca, Caruso, Augello, and Sabatucci}]{badagliacca2025graph}
Dario Badagliacca, Gabriele Caruso, Agnese Augello, and Luca Sabatucci. 2025.
\newblock Graph of goal-oriented thoughts: Design and implementation of llm agents.
\newblock In \emph{SOCIALIZE 2025, CEUR Workshop Proceedings}.

\bibitem[{Besta et~al.(2024)Besta, Blach, Kubicek, Gerstenberger, Podstawski, Gianinazzi, Gajda, Lehmann, Niewiadomski, Nyczyk et~al.}]{besta2024graph}
Maciej Besta, Nils Blach, Ales Kubicek, Robert Gerstenberger, Michal Podstawski, Lukas Gianinazzi, Joanna Gajda, Tomasz Lehmann, Hubert Niewiadomski, Piotr Nyczyk, and 1 others. 2024.
\newblock Graph of thoughts: Solving elaborate problems with large language models.
\newblock In \emph{Proceedings of the AAAI Conference on Artificial Intelligence}, volume~38, pages 17682--17690.

\bibitem[{Cai and Wang(2020)}]{cai2020note}
Chen Cai and Yusu Wang. 2020.
\newblock A note on over-smoothing for graph neural networks.
\newblock In \emph{ICML 2020 Workshop on Graph Representation Learning and Beyond}.

\bibitem[{Chan et~al.(2023)Chan, Chen, Su, Yu, Xue, Zhang, Fu, and Liu}]{chan2023chateval}
Chi-Min Chan, Weize Chen, Yusheng Su, Jianxuan Yu, Wei Xue, Shanghang Zhang, Jie Fu, and Zhiyuan Liu. 2023.
\newblock Chateval: Towards better llm-based evaluators through multi-agent debate.
\newblock In \emph{Twelfth International Conference on Learning Representations}.

\bibitem[{Chen et~al.(2025{\natexlab{a}})Chen, Wu, Zhang, Yang, Huang, Wang, Wang, and Wang}]{chen2025survey}
Ada Chen, Yongjiang Wu, Junyuan Zhang, Shu Yang, Jen-tse Huang, Kun Wang, Wenxuan Wang, and Shuai Wang. 2025{\natexlab{a}}.
\newblock A survey on the safety and security threats of computer-using agents: Jarvis or ultron?
\newblock \emph{arXiv preprint arXiv:2505.10924}.

\bibitem[{Chen et~al.(2020)Chen, Lin, Li, Li, Zhou, and Sun}]{chen2020measuring}
Deli Chen, Yankai Lin, Wei Li, Peng Li, Jie Zhou, and Xu~Sun. 2020.
\newblock Measuring and relieving the over-smoothing problem for graph neural networks from the topological view.
\newblock In \emph{Proceedings of the AAAI conference on artificial intelligence}, volume~34, pages 3438--3445.

\bibitem[{Chen et~al.(2023)Chen, Ji, Xu, and Zhao}]{chen2023consensusllm}
Huaben Chen, Wenkang Ji, Lufeng Xu, and Shiyu Zhao. 2023.
\newblock Multi-agent consensus seeking via large language models.
\newblock \emph{arXiv preprint arXiv:2310.20151}.

\bibitem[{Chen et~al.(2025{\natexlab{b}})Chen, Ding, Yu, Huang, Yang, and Zhang}]{chen2025scitoolagent}
Huajun Chen, Keyan Ding, Jing Yu, Junjie Huang, Yuchen Yang, and Qiang Zhang. 2025{\natexlab{b}}.
\newblock Scitoolagent: A knowledge graph-driven scientific agent for multi-tool integration.

\bibitem[{Chen et~al.(2025{\natexlab{c}})Chen, Tang, Deng, Wu, Wu, Jiang, Prasanna, Cohan, and Wang}]{chen2025locagent}
Zhaoling Chen, Xiangru Tang, Gangda Deng, Fang Wu, Jialong Wu, Zhiwei Jiang, Viktor Prasanna, Arman Cohan, and Xingyao Wang. 2025{\natexlab{c}}.
\newblock Locagent: Graph-guided llm agents for code localization.
\newblock In \emph{Proceedings of the 63rd Annual Meeting of the Association for Computational Linguistics}.

\bibitem[{Du et~al.(2024)Du, Li, Torralba, Tenenbaum, and Mordatch}]{du2023improving}
Yilun Du, Shuang Li, Antonio Torralba, Joshua~B Tenenbaum, and Igor Mordatch. 2024.
\newblock Improving factuality and reasoning in language models through multiagent debate.
\newblock In \emph{Proceedings of the 41st International Conference on Machine Learning}, pages 11733--11763.

\bibitem[{Eo et~al.(2025)Eo, Moon, Zi, Park, and Lim}]{eo2025debate-down}
Sugyeong Eo, Hyeonseok Moon, Evelyn~Hayoon Zi, Chanjun Park, and Heuiseok Lim. 2025.
\newblock Debate only when necessary: Adaptive multiagent collaboration for efficient llm reasoning.
\newblock \emph{arXiv preprint arXiv:2504.05047}.

\bibitem[{Fan et~al.(2024)Fan, Ding, Ning, Wang, Li, Yin, Chua, and Li}]{fan2024survey}
Wenqi Fan, Yujuan Ding, Liangbo Ning, Shijie Wang, Hengyun Li, Dawei Yin, Tat-Seng Chua, and Qing Li. 2024.
\newblock A survey on rag meeting llms: Towards retrieval-augmented large language models.
\newblock In \emph{Proceedings of the 30th ACM SIGKDD Conference on Knowledge Discovery and Data Mining}, pages 6491--6501.

\bibitem[{Gao et~al.(2025)Gao, Liu, He, Dou, Du, Deng, Hooi, Lin, and Pang}]{liuyue_FlowReasoner}
Hongcheng Gao, Yue Liu, Yufei He, Longxu Dou, Chao Du, Zhijie Deng, Bryan Hooi, Min Lin, and Tianyu Pang. 2025.
\newblock Flowreasoner: Reinforcing query-level meta-agents.
\newblock \emph{arXiv preprint arXiv:2504.15257}.

\bibitem[{Guo et~al.(2024)Guo, Chen, Wang, Chang, Pei, Chawla, Wiest, and Zhang}]{guo2024large}
Taicheng Guo, Xiuying Chen, Yaqi Wang, Ruidi Chang, Shichao Pei, Nitesh~V Chawla, Olaf Wiest, and Xiangliang Zhang. 2024.
\newblock Large language model based multi-agents: a survey of progress and challenges.
\newblock In \emph{Proceedings of the Thirty-Third International Joint Conference on Artificial Intelligence}, pages 8048--8057.

\bibitem[{Hsu et~al.(2024)Hsu, Cox, Xu, Tan, Zhai, Hu, Pratt, Chen, Hu, and Ding}]{hsu2024thought}
Chi-Yang Hsu, Kyle Cox, Jiawei Xu, Zhen Tan, Tianhua Zhai, Mengzhou Hu, Dexter Pratt, Tianlong Chen, Ziniu Hu, and Ying Ding. 2024.
\newblock Thought graph: Generating thought process for biological reasoning.
\newblock In \emph{Companion Proceedings of the ACM Web Conference 2024}, pages 537--540.

\bibitem[{Hu et~al.(2025)Hu, Cai, Du, Zhu, Liu, Yu, Hou, Tang, and Chen}]{hu2024self-evomac}
Yue Hu, Yuzhu Cai, Yaxin Du, Xinyu Zhu, Xiangrui Liu, Zijie Yu, Yuchen Hou, Shuo Tang, and Siheng Chen. 2025.
\newblock Self-evolving multi-agent collaboration networks for software development.
\newblock In \emph{Thirteenth International Conference on Learning Representations}.

\bibitem[{Huang et~al.(2025)Huang, Pan, and Ye}]{huang2025graphormer}
Wanjing Huang, Tongjie Pan, and Yalan Ye. 2025.
\newblock Graphormer-guided task planning: Beyond static rules with llm safety perception.
\newblock \emph{arXiv preprint arXiv:2503.06866}.

\bibitem[{Huang et~al.(2024)Huang, Liu, Chen, Wang, Wang, Lian, Wang, Tang, and Chen}]{huang2024understanding}
Xu~Huang, Weiwen Liu, Xiaolong Chen, Xingmei Wang, Hao Wang, Defu Lian, Yasheng Wang, Ruiming Tang, and Enhong Chen. 2024.
\newblock Understanding the planning of llm agents: A survey.
\newblock \emph{arXiv preprint arXiv:2402.02716}.

\bibitem[{Jiang et~al.(2025)Jiang, Zhou, Zhao, Song, Zhu, Zhu, and Wen}]{jiang2024kg}
Jinhao Jiang, Kun Zhou, Wayne~Xin Zhao, Yang Song, Chen Zhu, Hengshu Zhu, and Ji-Rong Wen. 2025.
\newblock Kg-agent: An efficient autonomous agent framework for complex reasoning over knowledge graph.
\newblock In \emph{Proceedings of the 63rd Annual Meeting of the Association for Computational Linguistics}.

\bibitem[{Jin et~al.(2020)Jin, Ma, Liu, Tang, Wang, and Tang}]{jin2020prognn}
Wei Jin, Yao Ma, Xiaorui Liu, Xianfeng Tang, Suhang Wang, and Jiliang Tang. 2020.
\newblock Graph structure learning for robust graph neural networks.
\newblock In \emph{Proceedings of the 26th ACM SIGKDD international conference on knowledge discovery \& data mining}, pages 66--74.

\bibitem[{Khan et~al.(2024)Khan, Hughes, Valentine, Ruis, Sachan, Radhakrishnan, Grefenstette, Bowman, Rockt{\"a}schel, and Perez}]{khan2024debating}
Akbir Khan, John Hughes, Dan Valentine, Laura Ruis, Kshitij Sachan, Ansh Radhakrishnan, Edward Grefenstette, Samuel~R Bowman, Tim Rockt{\"a}schel, and Ethan Perez. 2024.
\newblock Debating with more persuasive llms leads to more truthful answers.
\newblock In \emph{Proceedings of the 41st International Conference on Machine Learning}, pages 23662--23733.

\bibitem[{Li et~al.(2025{\natexlab{a}})Li, Liu, Wen, Zhang, and Pan}]{li2025assemble}
Shiyuan Li, Yixin Liu, Qingsong Wen, Chengqi Zhang, and Shirui Pan. 2025{\natexlab{a}}.
\newblock Assemble your crew: Automatic multi-agent communication topology design via autoregressive graph generation.
\newblock \emph{arXiv preprint arXiv:2507.18224}.

\bibitem[{Li et~al.(2024{\natexlab{a}})Li, Li, Wang, Li, Sun, Cheng, and Yu}]{li2023survey}
Yuhan Li, Zhixun Li, Peisong Wang, Jia Li, Xiangguo Sun, Hong Cheng, and Jeffrey~Xu Yu. 2024{\natexlab{a}}.
\newblock A survey of graph meets large language model: Progress and future directions.
\newblock In \emph{Proceedings of the Thirty-Third International Joint Conference on Artificial Intelligence}.

\bibitem[{Li et~al.(2024{\natexlab{b}})Li, Du, Zhang, Hou, Grabowski, Li, and Ie}]{li2024improving-sparse-mad}
Yunxuan Li, Yibing Du, Jiageng Zhang, Le~Hou, Peter Grabowski, Yeqing Li, and Eugene Ie. 2024{\natexlab{b}}.
\newblock Improving multi-agent debate with sparse communication topology.
\newblock In \emph{Findings of the Association for Computational Linguistics: EMNLP 2024}, pages 7281--7294.

\bibitem[{Li et~al.(2025{\natexlab{b}})Li, Mi, Zhou, Jiang, Zhang, Wang, and Fang}]{li2025goal}
Zherui Li, Yan Mi, Zhenhong Zhou, Houcheng Jiang, Guibin Zhang, Kun Wang, and Junfeng Fang. 2025{\natexlab{b}}.
\newblock Goal-aware identification and rectification of misinformation in multi-agent systems.
\newblock \emph{arXiv preprint arXiv:2506.00509}.

\bibitem[{Li et~al.(2025{\natexlab{c}})Li, Chen, Zhao, Wang, Liu, Zhang, Zhou, and Yu}]{li2025iceberg}
Zhixun Li, Dingshuo Chen, Tong Zhao, Daixin Wang, Hongrui Liu, Zhiqiang Zhang, Jun Zhou, and Jeffrey~Xu Yu. 2025{\natexlab{c}}.
\newblock Iceberg: Debiased self-training for class-imbalanced node classification.
\newblock In \emph{Proceedings of the ACM on Web Conference 2025}, pages 3160--3170.

\bibitem[{Li et~al.(2023)Li, Wang, Sun, Luo, Zhu, Chen, Luo, Zhou, Liu, Wu et~al.}]{li2023gslb}
Zhixun Li, Liang Wang, Xin Sun, Yifan Luo, Yanqiao Zhu, Dingshuo Chen, Yingtao Luo, Xiangxin Zhou, Qiang Liu, Shu Wu, and 1 others. 2023.
\newblock Gslb: The graph structure learning benchmark.
\newblock \emph{Advances in Neural Information Processing Systems}, 36:30306--30318.

\bibitem[{Liang et~al.(2024)Liang, He, Jiao, Wang, Wang, Wang, Yang, Shi, and Tu}]{liang2023encouraging}
Tian Liang, Zhiwei He, Wenxiang Jiao, Xing Wang, Yan Wang, Rui Wang, Yujiu Yang, Shuming Shi, and Zhaopeng Tu. 2024.
\newblock Encouraging divergent thinking in large language models through multi-agent debate.
\newblock In \emph{Proceedings of the 2024 Conference on Empirical Methods in Natural Language Processing}.

\bibitem[{Liu et~al.(2025{\natexlab{a}})Liu, Ying, Wang, Mu, Guo, Wang, Ma, Liang, Zhang, Liu et~al.}]{liu2025agentsafe}
Aishan Liu, Zonghao Ying, Le~Wang, Junjie Mu, Jinyang Guo, Jiakai Wang, Yuqing Ma, Siyuan Liang, Mingchuan Zhang, Xianglong Liu, and 1 others. 2025{\natexlab{a}}.
\newblock Agentsafe: Benchmarking the safety of embodied agents on hazardous instructions.
\newblock \emph{arXiv preprint arXiv:2506.14697}.

\bibitem[{Liu et~al.(2025{\natexlab{b}})Liu, Zhou, Liu, Zhang, Liang, Wang, Pu, Li, Zhang, Zhou et~al.}]{liu2025compromising}
Aishan Liu, Yuguang Zhou, Xianglong Liu, Tianyuan Zhang, Siyuan Liang, Jiakai Wang, Yanjun Pu, Tianlin Li, Junqi Zhang, Wenbo Zhou, and 1 others. 2025{\natexlab{b}}.
\newblock Compromising llm driven embodied agents with contextual backdoor attacks.
\newblock \emph{IEEE Transactions on Information Forensics and Security}.

\bibitem[{Liu et~al.(2024{\natexlab{a}})Liu, Feng, Xue, Wang, Wu, Lu, Zhao, Deng, Zhang, Ruan et~al.}]{liu2024deepseek}
Aixin Liu, Bei Feng, Bing Xue, Bingxuan Wang, Bochao Wu, Chengda Lu, Chenggang Zhao, Chengqi Deng, Chenyu Zhang, Chong Ruan, and 1 others. 2024{\natexlab{a}}.
\newblock {DeepSeek-v3} technical report.
\newblock \emph{arXiv preprint arXiv:2412.19437}.

\bibitem[{Liu et~al.(2024{\natexlab{b}})Liu, Peng, Yi, Xie, Xiang, Liu, and Xu}]{liu2024toolnet}
Xukun Liu, Zhiyuan Peng, Xiaoyuan Yi, Xing Xie, Lirong Xiang, Yuchen Liu, and Dongkuan Xu. 2024{\natexlab{b}}.
\newblock Toolnet: Connecting large language models with massive tools via tool graph.
\newblock \emph{arXiv preprint arXiv:2403.00839}.

\bibitem[{Liu et~al.(2024{\natexlab{c}})Liu, Lai, Gao, Cui, Li, Zhu, Lu, Chen, Qiao, Dai et~al.}]{liu2024controlllm}
Zhaoyang Liu, Zeqiang Lai, Zhangwei Gao, Erfei Cui, Ziheng Li, Xizhou Zhu, Lewei Lu, Qifeng Chen, Yu~Qiao, Jifeng Dai, and 1 others. 2024{\natexlab{c}}.
\newblock Controlllm: Augment language models with tools by searching on graphs.
\newblock In \emph{European Conference on Computer Vision}, pages 89--105.

\bibitem[{Liu et~al.(2023)Liu, Zhang, Li, Liu, and Yang}]{liu2023dylan}
Zijun Liu, Yanzhe Zhang, Peng Li, Yang Liu, and Diyi Yang. 2023.
\newblock Dynamic llm-agent network: An llm-agent collaboration framework with agent team optimization.
\newblock \emph{arXiv preprint arXiv:2310.02170}.

\bibitem[{Luo et~al.(2021)Luo, Cheng, Yu, Zong, Ni, Chen, and Zhang}]{luo2021learning}
Dongsheng Luo, Wei Cheng, Wenchao Yu, Bo~Zong, Jingchao Ni, Haifeng Chen, and Xiang Zhang. 2021.
\newblock Learning to drop: Robust graph neural network via topological denoising.
\newblock In \emph{Proceedings of the 14th ACM international conference on web search and data mining}, pages 779--787.

\bibitem[{Luo et~al.(2025{\natexlab{a}})Luo, Dai, Ni, Li, Zhang, Wang, Liu, and Salam}]{luo2025agentauditor}
Hanjun Luo, Shenyu Dai, Chiming Ni, Xinfeng Li, Guibin Zhang, Kun Wang, Tongliang Liu, and Hanan Salam. 2025{\natexlab{a}}.
\newblock Agentauditor: Human-level safety and security evaluation for llm agents.
\newblock \emph{arXiv preprint arXiv:2506.00641}.

\bibitem[{Luo et~al.(2025{\natexlab{b}})Luo, Zhao, Haffari, Phung, Gong, and Pan}]{luo2025gfm}
Linhao Luo, Zicheng Zhao, Gholamreza Haffari, Dinh Phung, Chen Gong, and Shirui Pan. 2025{\natexlab{b}}.
\newblock Gfm-rag: Graph foundation model for retrieval augmented generation.
\newblock \emph{arXiv preprint arXiv:2502.01113}.

\bibitem[{M.~Bran et~al.(2024)M.~Bran, Cox, Schilter, Baldassari, White, and Schwaller}]{m2024augmenting}
Andres M.~Bran, Sam Cox, Oliver Schilter, Carlo Baldassari, Andrew~D White, and Philippe Schwaller. 2024.
\newblock Augmenting large language models with chemistry tools.
\newblock \emph{Nature Machine Intelligence}, 6(5):525--535.

\bibitem[{Manish(2024)}]{manish2024autonomous}
Sanwal Manish. 2024.
\newblock An autonomous multi-agent llm framework for agile software development.
\newblock \emph{International Journal of Trend in Scientific Research and Development}, 8(5):892--898.

\bibitem[{Mao et~al.(2025)Mao, Meng, Duan, Yu, Jia, Fang, Liang, Wang, and Wen}]{mao2025agentsafe}
Junyuan Mao, Fanci Meng, Yifan Duan, Miao Yu, Xiaojun Jia, Junfeng Fang, Yuxuan Liang, Kun Wang, and Qingsong Wen. 2025.
\newblock Agentsafe: Safeguarding large language model-based multi-agent systems via hierarchical data management.
\newblock \emph{arXiv preprint arXiv:2503.04392}.

\bibitem[{Ning and Xie(2024)}]{ning2024survey}
Zepeng Ning and Lihua Xie. 2024.
\newblock A survey on multi-agent reinforcement learning and its application.
\newblock \emph{Journal of Automation and Intelligence}, 3(2):73--91.

\bibitem[{Papp et~al.(2021)Papp, Martinkus, Faber, and Wattenhofer}]{papp2021dropgnn}
P{\'a}l~Andr{\'a}s Papp, Karolis Martinkus, Lukas Faber, and Roger Wattenhofer. 2021.
\newblock Dropgnn: Random dropouts increase the expressiveness of graph neural networks.
\newblock \emph{Advances in Neural Information Processing Systems}, 34:21997--22009.

\bibitem[{Qian et~al.(2025)Qian, Xie, Wang, Liu, Dang, Du, Chen, Yang, Liu, and Sun}]{qian2024scaling-macnet}
Chen Qian, Zihao Xie, Yifei Wang, Wei Liu, Yufan Dang, Zhuoyun Du, Weize Chen, Cheng Yang, Zhiyuan Liu, and Maosong Sun. 2025.
\newblock Scaling large-language-model-based multi-agent collaboration.
\newblock In \emph{Thirteenth International Conference on Learning Representations}.

\bibitem[{Qiao et~al.(2025)Qiao, Fang, Qiu, Wang, Zhang, Jiang, Xie, Huang, and Chen}]{qiao2024benchmarking}
Shuofei Qiao, Runnan Fang, Zhisong Qiu, Xiaobin Wang, Ningyu Zhang, Yong Jiang, Pengjun Xie, Fei Huang, and Huajun Chen. 2025.
\newblock Benchmarking agentic workflow generation.
\newblock In \emph{Thirteenth International Conference on Learning Representations}.

\bibitem[{Rong et~al.(2020)Rong, Huang, Xu, and Huang}]{rong2019dropedge}
Yu~Rong, Wenbing Huang, Tingyang Xu, and Junzhou Huang. 2020.
\newblock Dropedge: Towards deep graph convolutional networks on node classification.
\newblock In \emph{Eight International Conference on Learning Representations}.

\bibitem[{Shen et~al.(2025)Shen, Liu, Dai, Wang, Miao, Tan, Pan, and Wang}]{shen2025understanding}
Xu~Shen, Yixin Liu, Yiwei Dai, Yili Wang, Rui Miao, Yue Tan, Shirui Pan, and Xin Wang. 2025.
\newblock Understanding the information propagation effects of communication topologies in llm-based multi-agent systems.
\newblock In \emph{Findings of the Association for Computational Linguistics: EMNLP 2025}.

\bibitem[{Shen et~al.(2023)Shen, Song, Tan, Li, Lu, and Zhuang}]{shen2023hugginggpt}
Yongliang Shen, Kaitao Song, Xu~Tan, Dongsheng Li, Weiming Lu, and Yueting Zhuang. 2023.
\newblock Hugginggpt: Solving ai tasks with chatgpt and its friends in hugging face.
\newblock \emph{Advances in Neural Information Processing Systems}, 36:38154--38180.

\bibitem[{Spielman and Srivastava(2008)}]{spielman2008graph-sparse-er}
Daniel~A Spielman and Nikhil Srivastava. 2008.
\newblock Graph sparsification by effective resistances.
\newblock In \emph{Proceedings of the fortieth annual ACM symposium on Theory of computing}, pages 563--568.

\bibitem[{Team et~al.(2024)Team, Georgiev, Lei, Burnell, Bai, Gulati, Tanzer, Vincent, Pan, Wang et~al.}]{team2024gemini}
Gemini Team, Petko Georgiev, Ving~Ian Lei, Ryan Burnell, Libin Bai, Anmol Gulati, Garrett Tanzer, Damien Vincent, Zhufeng Pan, Shibo Wang, and 1 others. 2024.
\newblock Gemini 1.5: Unlocking multimodal understanding across millions of tokens of context.
\newblock \emph{arXiv preprint arXiv:2403.05530}.

\bibitem[{Wang et~al.(2025{\natexlab{a}})Wang, Zhang, Zhou, Wu, Yu, Zhao, Yin, Fu, Yan, Luo et~al.}]{wang2025comprehensive}
Kun Wang, Guibin Zhang, Zhenhong Zhou, Jiahao Wu, Miao Yu, Shiqian Zhao, Chenlong Yin, Jinhu Fu, Yibo Yan, Hanjun Luo, and 1 others. 2025{\natexlab{a}}.
\newblock A comprehensive survey in llm (-agent) full stack safety: Data, training and deployment.
\newblock \emph{arXiv preprint arXiv:2504.15585}.

\bibitem[{Wang et~al.(2025{\natexlab{b}})Wang, Ying, Zhang, Liang, Hu, Zhang, Liu, and Liu}]{wang2025manipulating}
Le~Wang, Zonghao Ying, Tianyuan Zhang, Siyuan Liang, Shengshan Hu, Mingchuan Zhang, Aishan Liu, and Xianglong Liu. 2025{\natexlab{b}}.
\newblock Manipulating multimodal agents via cross-modal prompt injection.
\newblock \emph{arXiv preprint arXiv:2504.14348}.

\bibitem[{Wang et~al.(2024)Wang, Ma, Feng, Zhang, Yang, Zhang, Chen, Tang, Chen, Lin et~al.}]{wang2024survey}
Lei Wang, Chen Ma, Xueyang Feng, Zeyu Zhang, Hao Yang, Jingsen Zhang, Zhiyuan Chen, Jiakai Tang, Xu~Chen, Yankai Lin, and 1 others. 2024.
\newblock A survey on large language model based autonomous agents.
\newblock \emph{Frontiers of Computer Science}, 18(6):186345.

\bibitem[{Wang et~al.(2025{\natexlab{c}})Wang, Zhang, Yu, Wan, Meng, Guo, Wang, and Wang}]{wang2025g}
Shilong Wang, Guibin Zhang, Miao Yu, Guancheng Wan, Fanci Meng, Chongye Guo, Kun Wang, and Yang Wang. 2025{\natexlab{c}}.
\newblock G-safeguard: A topology-guided security lens and treatment on llm-based multi-agent systems.
\newblock In \emph{Proceedings of the 63rd Annual Meeting of the Association for Computational Linguistics}.

\bibitem[{Wang et~al.(2025{\natexlab{d}})Wang, Tan, Chen, Zhou, Chen, and Li}]{wang2025anymac}
Song Wang, Zhen Tan, Zihan Chen, Shuang Zhou, Tianlong Chen, and Jundong Li. 2025{\natexlab{d}}.
\newblock Anymac: Cascading flexible multi-agent collaboration via next-agent prediction.
\newblock \emph{arXiv preprint arXiv:2506.17784}.

\bibitem[{Wang et~al.(2025{\natexlab{e}})Wang, Zeng, Liu, Li, Wang, Shang, Jiang, Liu, and Wong}]{wang2024toolflow}
Zezhong Wang, Xingshan Zeng, Weiwen Liu, Liangyou Li, Yasheng Wang, Lifeng Shang, Xin Jiang, Qun Liu, and Kam-Fai Wong. 2025{\natexlab{e}}.
\newblock Toolflow: Boosting llm tool-calling through natural and coherent dialogue synthesis.
\newblock In \emph{Proceedings of the 2025 Conference of the Nations of the Americas Chapter of the Association for Computational Linguistics: Human Language Technologies (Volume 1: Long Papers)}, pages 4246--4263.

\bibitem[{Wang et~al.(2025{\natexlab{f}})Wang, Moriyama, Wang, Gangopadhyay, and Takamatsu}]{wang2025talk}
Zhao Wang, Sota Moriyama, Wei-Yao Wang, Briti Gangopadhyay, and Shingo Takamatsu. 2025{\natexlab{f}}.
\newblock Talk structurally, act hierarchically: A collaborative framework for llm multi-agent systems.
\newblock \emph{arXiv preprint arXiv:2502.11098}.

\bibitem[{Wang et~al.(2025{\natexlab{g}})Wang, Wang, Liu, Ding, Zhang, Liu, and Zhang}]{wang2025agentdropout}
Zhexuan Wang, Yutong Wang, Xuebo Liu, Liang Ding, Miao Zhang, Jie Liu, and Min Zhang. 2025{\natexlab{g}}.
\newblock Agentdropout: Dynamic agent elimination for token-efficient and high-performance llm-based multi-agent collaboration.
\newblock \emph{arXiv preprint arXiv:2503.18891}.

\bibitem[{Wei et~al.(2022)Wei, Wang, Schuurmans, Bosma, Xia, Chi, Le, Zhou et~al.}]{wei2022chain}
Jason Wei, Xuezhi Wang, Dale Schuurmans, Maarten Bosma, Fei Xia, Ed~Chi, Quoc~V Le, Denny Zhou, and 1 others. 2022.
\newblock Chain-of-thought prompting elicits reasoning in large language models.
\newblock \emph{Advances in neural information processing systems}, 35:24824--24837.

\bibitem[{Wei et~al.(2025)Wei, Huang, Li, Xing, Lin, and He}]{wei2025vflow}
Yangbo Wei, Zhen Huang, Huang Li, Wei~W Xing, Ting-Jung Lin, and Lei He. 2025.
\newblock Vflow: Discovering optimal agentic workflows for verilog generation.
\newblock \emph{arXiv preprint arXiv:2504.03723}.

\bibitem[{Wu et~al.(2024{\natexlab{a}})Wu, Bansal, Zhang, Wu, Li, Zhu, Jiang, Zhang, Zhang, Liu et~al.}]{wu2024autogen}
Qingyun Wu, Gagan Bansal, Jieyu Zhang, Yiran Wu, Beibin Li, Erkang Zhu, Li~Jiang, Xiaoyun Zhang, Shaokun Zhang, Jiale Liu, and 1 others. 2024{\natexlab{a}}.
\newblock Autogen: Enabling next-gen llm applications via multi-agent conversation.
\newblock In \emph{ICLR 2024 Workshop on Large Language Model (LLM) Agents}.

\bibitem[{Wu et~al.(2024{\natexlab{b}})Wu, Shen, Shan, Song, Wang, Zhang, Feng, Cheng, Chen, Xiong et~al.}]{wu2024can}
Xixi Wu, Yifei Shen, Caihua Shan, Kaitao Song, Siwei Wang, Bohang Zhang, Jiarui Feng, Hong Cheng, Wei Chen, Yun Xiong, and 1 others. 2024{\natexlab{b}}.
\newblock Can graph learning improve planning in llm-based agents?
\newblock In \emph{The Thirty-eighth Annual Conference on Neural Information Processing Systems}.

\bibitem[{Wu et~al.(2024{\natexlab{c}})Wu, Fan, Min, Prabhumoye, McAleer, Bisk, Salakhutdinov, Li, and Mitchell}]{wu2024agentkit}
Yue Wu, Yewen Fan, So~Yeon Min, Shrimai Prabhumoye, Stephen McAleer, Yonatan Bisk, Ruslan Salakhutdinov, Yuanzhi Li, and Tom Mitchell. 2024{\natexlab{c}}.
\newblock Agentkit: Structured llm reasoning with dynamic graphs.
\newblock In \emph{First Conference on Language Modeling}.

\bibitem[{Wu et~al.(2020)Wu, Pan, Chen, Long, Zhang, and Philip}]{wu2020comprehensive}
Zonghan Wu, Shirui Pan, Fengwen Chen, Guodong Long, Chengqi Zhang, and S~Yu Philip. 2020.
\newblock A comprehensive survey on graph neural networks.
\newblock \emph{IEEE transactions on neural networks and learning systems}, 32(1):4--24.

\bibitem[{Xi et~al.(2025)Xi, Chen, Guo, He, Ding, Hong, Zhang, Wang, Jin, Zhou et~al.}]{xi2025rise}
Zhiheng Xi, Wenxiang Chen, Xin Guo, Wei He, Yiwen Ding, Boyang Hong, Ming Zhang, Junzhe Wang, Senjie Jin, Enyu Zhou, and 1 others. 2025.
\newblock The rise and potential of large language model based agents: A survey.
\newblock \emph{Science China Information Sciences}, 68(2):121101.

\bibitem[{Xie et~al.(2025)Xie, Han, Shi, Cui, Zhao, Wu, and Zhao}]{xie2025rmoa}
Zhentao Xie, Chengcheng Han, Jinxin Shi, Wenjun Cui, Xin Zhao, Xingjiao Wu, and Jiabao Zhao. 2025.
\newblock Rmoa: Optimizing mixture-of-agents through diversity maximization and residual compensation.
\newblock \emph{arXiv preprint arXiv:2505.24442}.

\bibitem[{Xu et~al.(2025)Xu, Mei, Gao, Tan, Liang, and Zhang}]{xu2025mem}
Wujiang Xu, Kai Mei, Hang Gao, Juntao Tan, Zujie Liang, and Yongfeng Zhang. 2025.
\newblock A-mem: Agentic memory for llm agents.
\newblock \emph{arXiv preprint arXiv:2502.12110}.

\bibitem[{Yao et~al.(2023)Yao, Yu, Zhao, Shafran, Griffiths, Cao, and Narasimhan}]{yao2023tree}
Shunyu Yao, Dian Yu, Jeffrey Zhao, Izhak Shafran, Tom Griffiths, Yuan Cao, and Karthik Narasimhan. 2023.
\newblock Tree of thoughts: Deliberate problem solving with large language models.
\newblock \emph{Advances in neural information processing systems}, 36:11809--11822.

\bibitem[{Yu et~al.(2025{\natexlab{a}})Yu, Meng, Zhou, Wang, Mao, Pang, Chen, Wang, Li, Zhang et~al.}]{yu2025survey}
Miao Yu, Fanci Meng, Xinyun Zhou, Shilong Wang, Junyuan Mao, Linsey Pang, Tianlong Chen, Kun Wang, Xinfeng Li, Yongfeng Zhang, and 1 others. 2025{\natexlab{a}}.
\newblock A survey on trustworthy llm agents: Threats and countermeasures.
\newblock In \emph{Proceedings of the 31th ACM SIGKDD international conference on knowledge discovery \& data mining}.

\bibitem[{Yu et~al.(2025{\natexlab{b}})Yu, Wang, Zhang, Mao, Yin, Liu, Wen, Wang, and Wang}]{yu2024netsafe}
Miao Yu, Shilong Wang, Guibin Zhang, Junyuan Mao, Chenlong Yin, Qijiong Liu, Qingsong Wen, Kun Wang, and Yang Wang. 2025{\natexlab{b}}.
\newblock Netsafe: Exploring the topological safety of multi-agent networks.
\newblock In \emph{Proceedings of the 63rd Annual Meeting of the Association for Computational Linguistics}.

\bibitem[{Yu et~al.(2024)Yu, Yao, Li, Deng, Jiang, Cao, Chen, Suchow, Cui, Liu et~al.}]{yu2024fincon}
Yangyang Yu, Zhiyuan Yao, Haohang Li, Zhiyang Deng, Yuechen Jiang, Yupeng Cao, Zhi Chen, Jordan Suchow, Zhenyu Cui, Rong Liu, and 1 others. 2024.
\newblock Fincon: A synthesized llm multi-agent system with conceptual verbal reinforcement for enhanced financial decision making.
\newblock \emph{Advances in Neural Information Processing Systems}, 37:137010--137045.

\bibitem[{Zhai et~al.(2024)Zhai, Bai, Lin, Pan, Tong, Zhou, Suhr, Xie, LeCun, Ma et~al.}]{zhai2024fine}
Simon Zhai, Hao Bai, Zipeng Lin, Jiayi Pan, Peter Tong, Yifei Zhou, Alane Suhr, Saining Xie, Yann LeCun, Yi~Ma, and 1 others. 2024.
\newblock Fine-tuning large vision-language models as decision-making agents via reinforcement learning.
\newblock \emph{Advances in neural information processing systems}, 37:110935--110971.

\bibitem[{Zhang et~al.(2025{\natexlab{a}})Zhang, Yang, Liu, Li, Han, Chen, Huang, Fu, and Yu}]{zhang2025appagent}
Chi Zhang, Zhao Yang, Jiaxuan Liu, Yanda Li, Yucheng Han, Xin Chen, Zebiao Huang, Bin Fu, and Gang Yu. 2025{\natexlab{a}}.
\newblock Appagent: Multimodal agents as smartphone users.
\newblock In \emph{Proceedings of the 2025 CHI Conference on Human Factors in Computing Systems}, pages 1--20.

\bibitem[{Zhang et~al.(2025{\natexlab{b}})Zhang, Chen, Wan, Chang, Cheng, Wang, Hu, and Bai}]{zhang2025evoflow}
Guibin Zhang, Kaijie Chen, Guancheng Wan, Heng Chang, Hong Cheng, Kun Wang, Shuyue Hu, and Lei Bai. 2025{\natexlab{b}}.
\newblock Evoflow: Evolving diverse agentic workflows on the fly.
\newblock \emph{arXiv preprint arXiv:2502.07373}.

\bibitem[{Zhang et~al.(2025{\natexlab{c}})Zhang, Niu, Fang, Wang, Bai, and Wang}]{zhang2025maas}
Guibin Zhang, Luyang Niu, Junfeng Fang, Kun Wang, Lei Bai, and Xiang Wang. 2025{\natexlab{c}}.
\newblock Multi-agent architecture search via agentic supernet.
\newblock In \emph{Forty-second International Conference on Machine Learning}.

\bibitem[{Zhang et~al.(2025{\natexlab{d}})Zhang, Sun, Yue, Jiang, Wang, Chen, and Pan}]{zhang2024mog}
Guibin Zhang, Xiangguo Sun, Yanwei Yue, Chonghe Jiang, Kun Wang, Tianlong Chen, and Shirui Pan. 2025{\natexlab{d}}.
\newblock Graph sparsification via mixture of graphs.
\newblock In \emph{Thirteenth International Conference on Learning Representations}.

\bibitem[{Zhang et~al.(2025{\natexlab{e}})Zhang, Yue, Li, Yun, Wan, Wang, Cheng, Yu, and Chen}]{zhang2024agentprune}
Guibin Zhang, Yanwei Yue, Zhixun Li, Sukwon Yun, Guancheng Wan, Kun Wang, Dawei Cheng, Jeffrey~Xu Yu, and Tianlong Chen. 2025{\natexlab{e}}.
\newblock Cut the crap: An economical communication pipeline for llm-based multi-agent systems.
\newblock In \emph{Thirteenth International Conference on Learning Representations}.

\bibitem[{Zhang et~al.(2025{\natexlab{f}})Zhang, Yue, Sun, Wan, Yu, Fang, Wang, Chen, and Cheng}]{zhang2025g}
Guibin Zhang, Yanwei Yue, Xiangguo Sun, Guancheng Wan, Miao Yu, Junfeng Fang, Kun Wang, Tianlong Chen, and Dawei Cheng. 2025{\natexlab{f}}.
\newblock G-designer: Architecting multi-agent communication topologies via graph neural networks.
\newblock In \emph{Forty-second International Conference on Machine Learning}.

\bibitem[{Zhang et~al.(2024{\natexlab{a}})Zhang, Yue, Wang, Fang, Sui, Wang, Liang, Cheng, Pan, and Chen}]{zhang2024gst}
Guibin Zhang, Yanwei Yue, Kun Wang, Junfeng Fang, Yongduo Sui, Kai Wang, Yuxuan Liang, Dawei Cheng, Shirui Pan, and Tianlong Chen. 2024{\natexlab{a}}.
\newblock Two heads are better than one: boosting graph sparse training via semantic and topological awareness.
\newblock In \emph{Proceedings of the 41st International Conference on Machine Learning}, pages 60197--60219.

\bibitem[{Zhang et~al.(2025{\natexlab{g}})Zhang, Xiang, Yu, Teng, Chen, Chen, Zhuge, Cheng, Hong, Wang et~al.}]{zhang2024aflow}
Jiayi Zhang, Jinyu Xiang, Zhaoyang Yu, Fengwei Teng, Xionghui Chen, Jiaqi Chen, Mingchen Zhuge, Xin Cheng, Sirui Hong, Jinlin Wang, and 1 others. 2025{\natexlab{g}}.
\newblock Aflow: Automating agentic workflow generation.
\newblock In \emph{Thirteenth International Conference on Learning Representations}.

\bibitem[{Zhang et~al.(2025{\natexlab{h}})Zhang, Wang, Ren, Jiang, Wang, and Liu}]{zhang2025ratt}
Jinghan Zhang, Xiting Wang, Weijieying Ren, Lu~Jiang, Dongjie Wang, and Kunpeng Liu. 2025{\natexlab{h}}.
\newblock Ratt: A thought structure for coherent and correct llm reasoning.
\newblock In \emph{Proceedings of the AAAI Conference on Artificial Intelligence}, volume~39, pages 26733--26741.

\bibitem[{Zhang et~al.(2024{\natexlab{b}})Zhang, Xu, Zhang, Liu, Hooi, and Deng}]{zhang2023machinesom}
Jintian Zhang, Xin Xu, Ningyu Zhang, Ruibo Liu, Bryan Hooi, and Shumin Deng. 2024{\natexlab{b}}.
\newblock Exploring collaboration mechanisms for llm agents: A social psychology view.
\newblock In \emph{Proceedings of the 62nd Annual Meeting of the Association for Computational Linguistics}.

\bibitem[{Zhang et~al.(2025{\natexlab{i}})Zhang, Ma, Cao, Zhang, and Zhao}]{zhang2025plan}
Shiqi Zhang, Xinbei Ma, Zouying Cao, Zhuosheng Zhang, and Hai Zhao. 2025{\natexlab{i}}.
\newblock Plan-over-graph: Towards parallelable llm agent schedule.
\newblock \emph{arXiv preprint arXiv:2502.14563}.

\bibitem[{Zhang et~al.(2025{\natexlab{j}})Zhang, Hou, Tang, Chen, Zhang, Dong, and Chen}]{zhang2025gnns-predict-mas}
Yuanshuo Zhang, Yuchen Hou, Bohan Tang, Shuo Chen, Muhan Zhang, Xiaowen Dong, and Siheng Chen. 2025{\natexlab{j}}.
\newblock Gnns as predictors of agentic workflow performances.
\newblock \emph{arXiv preprint arXiv:2503.11301}.

\bibitem[{Zhang et~al.(2024{\natexlab{c}})Zhang, Cui, Lu, Zhou, Yang, Wang, and Huang}]{zhang2024agent}
Zhexin Zhang, Shiyao Cui, Yida Lu, Jingzhuo Zhou, Junxiao Yang, Hongning Wang, and Minlie Huang. 2024{\natexlab{c}}.
\newblock Agent-safetybench: Evaluating the safety of llm agents.
\newblock \emph{arXiv preprint arXiv:2412.14470}.

\bibitem[{Zhou et~al.(2025{\natexlab{a}})Zhou, Wan, Sun, Palangi, Iqbal, Vuli{\'c}, Korhonen, and Ar{\i}k}]{zhou2025multi}
Han Zhou, Xingchen Wan, Ruoxi Sun, Hamid Palangi, Shariq Iqbal, Ivan Vuli{\'c}, Anna Korhonen, and Sercan~{\"O} Ar{\i}k. 2025{\natexlab{a}}.
\newblock Multi-agent design: Optimizing agents with better prompts and topologies.
\newblock \emph{arXiv preprint arXiv:2502.02533}.

\bibitem[{Zhou et~al.(2025{\natexlab{b}})Zhou, Geng, Xue, Kang, Qin, Wang, Yin, and Bai}]{zhou2025reso}
Heng Zhou, Hejia Geng, Xiangyuan Xue, Li~Kang, Yiran Qin, Zhiyong Wang, Zhenfei Yin, and Lei Bai. 2025{\natexlab{b}}.
\newblock Reso: A reward-driven self-organizing llm-based multi-agent system for reasoning tasks.
\newblock \emph{arXiv preprint arXiv:2503.02390}.

\bibitem[{Zhou et~al.(2024)Zhou, Fan, Liu, Xu, Jin, and Li}]{zhou2024synergizing}
Zhilun Zhou, Jingyang Fan, Yu~Liu, Fengli Xu, Depeng Jin, and Yong Li. 2024.
\newblock Synergizing llm agents and knowledge graph for socioeconomic prediction in lbsn.
\newblock \emph{arXiv preprint arXiv:2411.00028}.

\bibitem[{Zhu et~al.(2024)Zhu, Dastani, and Wang}]{zhu2024survey}
Changxi Zhu, Mehdi Dastani, and Shihan Wang. 2024.
\newblock A survey of multi-agent deep reinforcement learning with communication.
\newblock \emph{Autonomous Agents and Multi-Agent Systems}, 38(1):4.

\bibitem[{Zhu et~al.(2021)Zhu, Xu, Zhang, Du, Zhang, Liu, Yang, and Wu}]{zhu2021survey-gsl}
Yanqiao Zhu, Weizhi Xu, Jinghao Zhang, Yuanqi Du, Jieyu Zhang, Qiang Liu, Carl Yang, and Shu Wu. 2021.
\newblock A survey on graph structure learning: Progress and opportunities.
\newblock \emph{arXiv preprint arXiv:2103.03036}.

\bibitem[{Zhuge et~al.(2024)Zhuge, Wang, Kirsch, Faccio, Khizbullin, and Schmidhuber}]{zhuge2024gptswarm}
Mingchen Zhuge, Wenyi Wang, Louis Kirsch, Francesco Faccio, Dmitrii Khizbullin, and J{\"u}rgen Schmidhuber. 2024.
\newblock Gptswarm: Language agents as optimizable graphs.
\newblock In \emph{Forty-first International Conference on Machine Learning}.

\end{thebibliography}

\end{document}